\relax
\documentclass[letterpaper]{article} 
\usepackage{aaai20}  
\usepackage{times}  
\usepackage{helvet} 
\usepackage{courier}  
\usepackage[hyphens]{url}  
\usepackage{graphicx} 
\usepackage{graphicx}  
\usepackage{amsmath}
\usepackage{multirow}
\usepackage{makecell}
\newcommand{\citet}[1]{\citeauthor{#1} \shortcite{#1}}
\newcommand{\citep}{\cite}

\title{DeGAN : Data-Enriching GAN for Retrieving Representative Samples \\from a Trained Classifier}

\author{Sravanti Addepalli, Gaurav Kumar Nayak, \\ \Large \textbf{Anirban Chakraborty, 
R. Venkatesh Babu}
\\
Department of Computational and Data Sciences\\
Indian Institute of Science, Bangalore, India\\
{ \{sravantia, gauravnayak, anirban, venky\}@iisc.ac.in}}
\begin{document}
 
\maketitle

\begin{abstract}
In this era of digital information explosion, an abundance of data from numerous modalities is being generated as well as archived everyday. However, most problems associated with training Deep Neural Networks still revolve around lack of data that is rich enough for a given task. Data is required not only for training an initial model, but also for future learning tasks such as Model Compression and Incremental Learning. A diverse dataset may be used for training an initial model, but it may not be feasible to store it throughout the product life cycle due to data privacy issues or memory constraints. We propose to bridge the gap between the abundance of available data and lack of relevant data, for the future learning tasks of a given trained network. We use the available data, that may be an imbalanced subset of the original training dataset, or a related domain dataset, to retrieve representative samples from a trained classifier, using a novel \textit{Data-enriching GAN} (DeGAN) framework. We demonstrate that data from a related domain can be leveraged to achieve state-of-the-art performance for the tasks of Data-free Knowledge Distillation and Incremental Learning on benchmark datasets. We further demonstrate that our proposed framework can enrich any data, even from unrelated domains, to make it more useful for the future learning tasks of a given network.
\end{abstract}

\section{Introduction}
\label{intro}
The performance and generalizability of Deep Neural Networks largely depend on the amount and quality of training data available. Several successful implementations of tasks such as classification, object detection and segmentation leverage very large, class-balanced and diverse datasets. In addition to training an initial network, data is also required for future updates to the model. This makes it important for training data to be available throughout the life cycle of a product. While data collection is a challenge in itself, storing the data for future use could also be a concern due to data confidentiality constraints, privacy issues, or memory costs. Business establishments may also protect their data to gain market advantage. There are several corporates that provide trained models to clients while maintaining the confidentiality of their data. The client may want to compress the model \cite{hinton2015distilling,srinivasdata} before deployment, expand the functionality of the model \cite{rebuffi2017icarl}, or refine the model based on domain specific data \cite{csurka2017domain}. These tasks typically require access to the initial training data, which may not be available to the client due to its proprietary nature. In addition, storage and future use of personal data could be limited due to privacy restrictions. This could include user details, biometric data or medical history.

Non-availability of data restricts future enhancements to a trained model. This issue has fuelled research in limited-data and data-free learning approaches to specific tasks such as Knowledge Distillation \cite{dfkd-nips-lld-17}, Incremental Learning \cite{castro2018end} and improving model robustness \cite{mopuri2018ask}. The key concern with Data-free approaches is that they operate in a severely constrained setting, where they assume non-availability of any additional data. This often leads to the process of reconstructing samples using variants of activation maximization \cite{erhan2009visualizing}, which is computationally expensive. Several iterations of back-propagation are required to generate one batch of representative samples.  One of the  concerns with few-shot learning approaches is that they require careful selection of exemplars \cite{castro2018end}, which may not be permitted in a privacy restricted setting. We aim to bridge the gap between data-free approaches, which over-constrain the problem setting; few-shot learning approaches, which assume availability of cherry-picked samples; and the traditional learning methods, that assume the entire training dataset to be available; by using related domain data as a proxy to the true data. 

While the lack of a rich, diverse dataset that is relevant to a given application is a key challenge, there are ever increasing sources of diverse data available on the web, which can potentially be tapped for the same. Such data however cannot be used directly as they may not belong to the same distribution as the training dataset, may not be diverse enough, and may not be equally represented across all classes. In this paper, we propose a Data-enriching GAN (DeGAN) framework to enrich \textit{any available data} to make it more useful for the future learning tasks of a pre-trained classifier. We term the \textit{any available data} used as \textit{Proxy Data}, or \textit{Proxy Dataset} as it serves as a proxy to the \textit{True dataset}. \textit{Proxy dataset} could comprise of unlabeled test data collected over a limited duration, or open source datasets, or a collection of images from the web, or synthetic images. In most practical industry applications, if a trained model is being enhanced for future use, it will have access to unlabeled test data. However, this test data may not be diverse enough, and it may contain data only from a few classes. Using such data directly for tasks such as Knowledge Distillation would lead to very poor performance. DeGAN can enrich this data to make it more representative of the training data, and introduce the diversity that is crucial for future learning tasks. 

As an example, a real-life use case of cancer screening is considered here, where an initial teacher model is trained using a large corpus of CT-scans of patients across various geographies. Training data is rich in terms of diversity and has class balance. However, the sensitivity and size of this data forbids its storage for future use. The teacher model is deployed for a few years, after which there is a requirement of deploying it on a handheld device with lower memory and compute. The organization may decide to use one month data from a given hospital to train the student net. This data is very much related to the true dataset, however it lacks the richness and diversity. It is also possibly class-imbalanced (containing non-cancerous classes only), leading to a degraded performance of the distilled network. Our proposed DeGAN is not only capable of generating a diverse set of samples, but can also handle the class-imbalance problem by using only one class data to generate representative samples for all classes.

In some applications such as Class-Incremental Learning, the availability of \textit{Proxy Dataset} is not an additional requirement. Here, an initial model is trained on old classes, which is incrementally trained on new class data in future. In a data-free scenario, old class data is assumed to be unavailable. Here, DeGAN can use the new class data as \textit{Proxy Dataset} to retrieve representative samples related to old classes from the pre-trained network.

The organization of this paper is as follows: The subsequent section outlines our contribution in this paper. This is followed by a discussion on the existing literature related to our work. Section \ref{prop_ap} gives a detailed description of our proposed approach. Following this, we present our experiments and results in Section \ref{exp}. We conclude the paper with our analysis of the proposed method in Section \ref{conc}.

\section{Contributions}

In this work, we propose a novel approach to retrieve representative samples from a pre-trained classifier using a three-player adversarial framework. We use a \textit{Proxy Dataset}, which is data from a related domain, but may be class-imbalanced, or composed of partially overlapping/ non-overlapping classes, as an aid to retrieve these samples. The three players in our proposed architecture are generator, discriminator and pre-trained classifier. In addition to the adversarial game between generator and discriminator that exists in a conventional GAN \cite{goodfellow2014generative} setup, we introduce an adversarial play between the discriminator and classifier as well.  While discriminator tries to bring the distribution of the generated data closer to that of the related domain (\textit{Proxy Dataset}), classifier tries to bring in features specific to the original training data distribution (\textit{True Dataset}). The classifier also ensures class balance in the generated samples. The result of this three-way adversarial training is that the generated samples lie on the image manifold of the \textit{Proxy Dataset}, while they also incorporate features from the \textit{True Dataset}. 

We consider the task of Knowledge Distillation to demonstrate that data from a related domain can be leveraged to achieve state-of-the-art performance for the future learning tasks of a pre-trained network. The process of generating samples is agnostic to any future task where they would potentially be used. This allows the generated data to be used for multiple tasks. We demonstrate proof-of-concept for this claim by considering the task of single-step Class-Incremental learning for CIFAR-$100$ dataset.
Our contribution in this work can be summarized as follows:
\begin{itemize}
	\item We propose a \textit{Data-enriching GAN} (DeGAN) framework to retrieve representative samples from a trained classifier using data from a related domain.
	\item We demonstrate state-of-the-art performance on the task of Data-Free Knowledge Distillation on CIFAR-$10$ \cite{krizhevsky2009learning} and Fashion MNIST \cite{xiao2017fashion} datasets using data generated by DeGAN. 
	\item We are the first to show results for Data-Free Knowledge Distillation on a dataset of larger size (CIFAR-$100$). This demonstrates the scalability of our data generation approach, when compared to the existing methods.
	\item We show that the proposed DeGAN could enrich data even from an unrelated domain, to make it more useful for the future learning tasks of a given network (such as using SVHN dataset for retrieving data from a CIFAR-$10$ classifier)
	\item We demonstrate state-of-the-art performance for the task of Data-free single-step Class Incremental Learning as a proof of concept of applicability of DeGAN to multiple Machine Learning tasks.
\end{itemize}
\section{Related Works}
\label{rel_work}
There are several works in existing literature, related to extracting representative samples from a trained classifier for various applications. \citet{simonyan2013deep} retrieve class specific samples from a deep convolutional network by maximizing their class scores, for the purpose of visualization. Similar ideas have been used by \citet{mopuri2018ask} for creating samples that are representative of a class, and using them for the task of crafting adversarial perturbations. These methods are computationally expensive as several iterations of back propagation are required to construct a single sample. Also, the generated samples are usually specific to a given task. For example, they may not be diverse enough to train a neural network and get optimal performance. Our proposed approach is task agnostic and is capable of generating samples that are diverse enough for tasks such as Knowledge Distillation and Class-Incremental Learning.
\subsection{Knowledge Distillation}
Knowledge distillation is a technique of transferring knowledge from a large capacity network (\textit{Teacher}) to a smaller network (\textit{Student}) without significant impact on accuracy. Early methods \cite{hinton2015distilling} utilize the entire training data for the task of distillation. \citet{li2018fewsamples} use $1$\% of the training data to train the \textit{Student} model. \citet{kimura2018few} create pseudo samples and augment them with few samples of the training data to train the \textit{Student} network. The pseudo samples are updated by increasing \textit{Student-Teacher} loss, whereas the network parameters are learned to reduce the same loss through an iterative and complicated optimization process. Here, the process of generating representative samples is task specific, as opposed to our proposed approach.

\citet{dfkd-nips-lld-17} use metadata to perform Knowledge Distillation. The statistics of training data are saved at each layer in the form of activation records, which are utilized to reconstruct the training samples. The work by \citet{nayak2019distillation} is an attempt towards Data-Free Knowledge Distillation, where access to metadata is also not required. The representative samples named \textit{Data Impressions}, are synthesized using the \textit{Teacher} model. Target vectors at the output softmax layer of the \textit{Teacher} network are sampled from a mixture of Dirichlet distributions with carefully selected parameters to ensure diversity. These samples are used to generate images such that the cross-entropy loss between the sampled vector and output of the network (corresponding to the generated images) is minimized. The images generated using this method are used for the task of Knowledge Distillation. Both these approaches require careful selection of parameters and are computationally expensive as multiple iterations of back-propagation are required to generate a single image.
\subsection{Class-Incremental Learning}
Incremental learning refers to the paradigm of learning continually from a stream of data. In Class-Incremental Learning \cite{rebuffi2017icarl}, a network is initially trained on a few classes, and is incrementally updated over time to learn new classes. During these updates, the deep model suffers from catastrophic forgetting \cite{catastrophic} as it forgets the mapping on old class data when only new class data is used to train the model. While a straight-forward method to overcome this issue is to simultaneously train the model using old class data and new class data, this trivial solution is not permitted, as it is assumed to be infeasible to store the old class data due to memory constraints, or other issues discussed earlier in Section \ref{intro}. This led to using methods such as \textit{finetuning}, which would not allow the model to be updated too much. Another baseline method is \textit{fixed representation}, where the parameters related to old classes are frozen and only the new class parameters are learned using the new data. These methods suffer from low accuracy either on old classes or new classes.

In order to avoid this, \citet{rebuffi2017icarl} choose a fixed number of exemplars from old class data based on a selection strategy called \textit{herding}, and use them along with new class data to train the incremental model. Cross entropy loss is used on the new class data, whereas distillation loss is used to retain performance on the old classes. \citet{castro2018end} improve the incremental model by jointly learning the data representation and classifier in an end to end fashion. Both these approaches use few exemplars from old class data to avoid catastrophic forgetting, which may not be feasible in a privacy constrained setting.

While \citet{shin2017continual} do not use old data for each incremental step, they train a generator using the old data. Our approach has broader applicability as we do not assume the availability of such a generator. \citet{li2017learning} assume a Data-Free setting, where there is no access to the old data. The new class samples are used to compute cross entropy loss on new classes, and distillation loss on old classes. While this approach can work well if the new class data is well distributed across all the old classes in the initial Classifier, it would not work in a case where the new class data is highly correlated to a small subset of the old classes. In such a case, the proposed DeGAN can be used to generate class-balanced representative samples of old classes using the new class samples as \textit{Proxy Data}. We demonstrate improved performance as compared to their results, which are reported by \citet{rebuffi2017icarl} for the Class-Incremental learning task on CIFAR-$100$ dataset.

We describe our proposed approach in detail in the following section.

\begin{figure*}[t]
\centering
\includegraphics[width=1\textwidth]{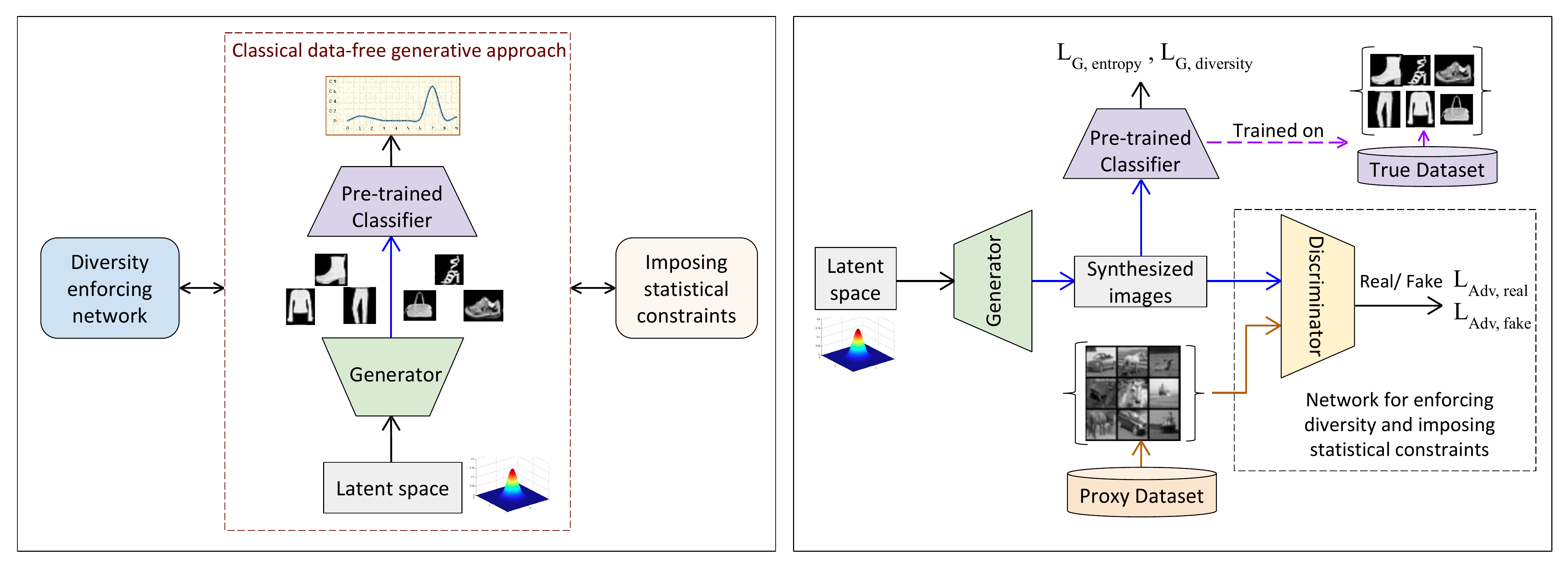}
\caption{(a) Block diagram showing the classical data-free generative approach with required additional constraints (b) Block diagram of the proposed DeGAN architecture}
\label{fig:architecture}
\end{figure*}

\section{Proposed Approach}
\label{prop_ap}
In this section, we first present a classical generative framework for retrieving representative samples from a trained classifier and discuss the associated issues. We further propose constraints that can be imposed to address these issues and discuss existing methods of imposing such constraints. We discuss the benefit of our proposal over the other implementations, followed by a detailed discussion on our proposed DeGAN framework.

\subsection{Data-Free Generative approach}
The central pathway of Fig. \ref{fig:architecture}(a) shows a classical data-free approach to generating samples using a generator and a pre-trained classifier. Inputs sampled from a latent space are utilised by the generator to produce images which are validated by the classifier. Some of the advantages of such a generative approach with respect to the conventional method of generating samples using activation maximization are, computational efficiency, memory efficiency and better diversity in generated data. However, using a generator could potentially lead to the following issues: mode collapse, and generation of noisy images that do not belong to the data distribution. \citet{nikolaidis2019learning} improve the diversity of images by training multiple generators and using all of them for the future goal of Knowledge Distillation. However, this process is inefficient and computationally expensive. Fig. \ref{fig:architecture}(a) illustrates the classical data free generative approach with the required additional constraints. In order to improve diversity, the architecture could include a \textit{diversity enforcing network} whose role is to build a one-to-one mapping from the output space of the generator to the input space. 

Since classifier is a many-to-one mapping function, the classical method of retrieving an input based on maximization of output activations can potentially lead to generation of images that are far from the true data distribution. Imposing additional statistical priors on the generated images can move the distribution of same closer to that of the true data. This process is closely related to enforcing visual quality on the generated images and has been well explored in the vision community \cite{weiss2007makes,zeng2017statistics,moorthy2011blind}. However, the constraints to be imposed are very specific to the dataset being considered. The process of hand-crafting these constraints can be tedious and assumes a lot of prior knowledge about the original training dataset. 

While one can use independent networks or loss functions to impose the constraints discussed above, this would lead to an increase in complexity. We can enforce the same constraints by intelligently utilising a single network (discriminator) in the framework, which motivates the proposed DeGAN, as explained in the following sections.

\subsection{Data-Enriching GAN (DeGAN)}
Designing novel metrics to impose visual priors individually on each dataset is a challenging task. Generative Adversarial Networks (GANs) \cite{goodfellow2014generative} are known to be useful for imposing priors on images for various tasks in image processing and computer vision. Hence, we introduce a discriminator that serves as the \textit{Imposing statistical constraints} block in Fig. \ref{fig:architecture}(a). Since we assume that the original training data is not available, we use data from a related domain (\textit{Proxy Data}) to train the generator-discriminator pair in an adversarial manner. The rationale behind this is that low level statistics of images remain same or similar for data from the same or related domains. Hence, GAN training ensures that the generated images lie on the manifold of the \textit{Proxy Data}, which is similar to that of the \textit{True Data}. 
There has been significant progress in the training methods and architectures of GANs to ensure diversity in the generated images \cite{radford2015unsupervised,salimans2016improved}. We leverage the progress in research on this front to address the issue of lack of diversity in the generated samples. Hence, the discriminator also serves the purpose of the \textit{Diversity enforcing network} in Fig. \ref{fig:architecture}(a). We use a Deep Convolutional GAN (DCGAN) \cite{radford2015unsupervised} for the experiments in this paper as it is very stable to train and scalable. Our proposed method can be used with other GAN architectures as well, and hence can be adapted to various applications.

Using a GAN enables us to generate data that belongs to the distribution of the \textit{Proxy Dataset}. However, we need to ensure that the learned distribution is close to the \textit{True Data} distribution. Therefore, we propose to use a three-player Data-enriching GAN for generating representative samples from a trained classifier. This consists of a Generator, Discriminator and Classifier coupled together as shown in Fig. \ref{fig:architecture}(b). Generator takes a multidimensional random vector as input, with each dimension sampled independently from a standard normal distribution. It generates an image which goes as input to the discriminator and classifier. Here, weights of the generator and discriminator are trainable, while weights of the classifier are frozen. The discriminator ensures that distribution of the generated data is close to that of the \textit{Proxy Dataset}. The role of the classifier is to ensure that the generated data incorporates features that the classifier expects in the input images. Classifier also ensures that the distribution of generated images is balanced across all the classes. 
\subsubsection{Loss Formulation:}
We consider $p_{\textit{\textbf{z}}}(\textit{\textbf{z}})$ as the distribution of the latent space (input space of the Generator), $\textit{\textbf{z}}$ to be a random vector sampled from $p_{\textit{\textbf{z}}}(\textit{\textbf{z}})$, $p_{data}(\textit{\textbf{x}})$ as the distribution of \textit{True Data}, $N$ as the number of images per batch, $K$ as the number of classes in the \textit{True Dataset}, $\lambda_{e}$ and $\lambda_{d}$ as positive constants which can be tuned to adjust the weightage of Entropy Loss and Diversity Loss respectively. We denote the Generator as $G$, Classifier as $C$, and Discriminator as $D$ here. We consider $\textit{\textbf{y}}$ as the Classifier output corresponding to the Generator input $\textit{\textbf{z}}$. The expectation over Classifier outputs across a batch of samples sampled from the latent space ($p_{\textit{\textbf{z}}}(\textit{\textbf{z}})$) is denoted by $\textit{\textbf{w}}$. 
\begin{flalign}
    & \textit{\textbf{y}} = C(G(\textit{\textbf{z}})), \;\;\textit{\textbf{w}} = E_{\textit{\textbf{z}}\sim p_{\textit{\textbf{z}}}(\textit{\textbf{z}})}[C(G(\textit{\textbf{z}}))] &
\end{flalign}
The losses used to train the proposed DeGAN are presented below:
\begin{itemize}
	\item Adversarial loss \cite{goodfellow2014generative} ($L_{Adv,real}$ and $L_{Adv,fake}$), to ensure that the distribution of the generated images is close to that of the \textit{Proxy Dataset},
	\begin{flalign}
    & L_{Adv, real} = E_{\textit{\textbf{x}} \sim p_{data}(\textit{\textbf{x}})}[log\; D(\textit{\textbf{x}})] & 
    \end{flalign}
    \begin{flalign}
    & L_{Adv, fake} = E_{\textit{\textbf{z}}\sim p_{\textit{\textbf{z}}}(\textit{\textbf{z}})}[log(1-D(G(\textit{\textbf{z}})))] & 
    \end{flalign}
	\item Entropy loss ($L_{G, entropy}$) (minimization of entropy of individual samples) at the output of the classifier, to ensure that each generated sample belongs to one of the classifier’s classes with high confidence,
    \begin{flalign}
    & L_{G, entropy} =  E_{\textit{\textbf{z}}\sim p_{\textit{\textbf{z}}}(\textit{\textbf{z}})}[-\sum_{k=0}^{K} y_{k} \;log(y_{k})] & 
    \end{flalign}
	\item Diversity loss ($L_{G, diversity}$) at the output of the classifier, to ensure that the entropy of the expected output of the classifier across a batch is high. When the individual outputs of the classifier are peaky, this loss ensures that the distribution of samples in a batch is uniform across classes. This prevents the generated samples from being biased towards any particular class.
	\begin{flalign}
    & L_{G, diversity} = -\sum_{k=0}^{K}w_{k}\; log(w_{k}) &
    \end{flalign}
\end{itemize}
 The equations below describe the Discriminator loss ($L_{D}$) and the Generator Loss ($L_{G}$). The Generator loss is composed of two additional losses imposed by the Classifier.
\begin{flalign}
    & L_{D} = L_{Adv, real} + L_{Adv, fake} & 
\end{flalign}
\begin{flalign}
    & L_{G} = L_{Adv, fake} + \lambda_{e}L_{G, entropy} - \lambda_{d}L_{G, diversity} &
\end{flalign}

We alternately maximize $L_{D}$ (freezing generator parameters) and minimize $L_{G}$ (freezing discriminator
parameters) in a manner similar to the standard GAN training \cite{goodfellow2014generative}. 
A combination of the above losses ensures that the generated images are similar to the related domain, incorporate features from the \textit{True Data} distribution and are equally distributed across all classes of the \textit{True Dataset}. 

In cases where the domain of the reference dataset is close to that of the \textit{True Dataset}, the value of $\lambda_{e}$ can be low (or set to $0$), as the images will already contain some of the features that the classifier expects. However, in cases where the reference dataset is not closely related to the \textit{True Dataset}, we need to give this loss a higher weightage. Class distribution and confidence of the generated images provide cues for tuning these hyperparameters. 

In the following section, we demonstrate that data retrieved using our proposed Data-enriching GAN can achieve state-of-the-art performance on the tasks of Knowledge Distillation and Class-Incremental Learning for some of the benchmark datasets. We also demonstrate scalability of our approach to CIFAR-$100$, which has not been done in any of the existing works. 
\begin{table}
  \caption{Accuracy (in \%) of Student network using Knowledge Distillation: Comparison with the state of the art ($^\#90$ non-overlapping classes from CIFAR-$100$ are used)\\}
  \label{cifar10-table}
  \centering
\resizebox{1\columnwidth}{!}{
  \begin{tabular}{l|ccc}
   \hline
    \textbf{True Dataset} & \textbf{CIFAR-10} & \textbf{F-MNIST} & \textbf{CIFAR-100} \\
    Proxy Dataset & CIFAR-$100^\#$  & CIFAR-$10$ & CIFAR-$10$ \\
    \hline
    \textit{Teacher} accuracy & 83.02 & 90.72 & 79.05\\
    Using \textit{True Data} & 81.78 & 88.98 & 69.65\\
    \citeauthor{kimura2018few}  & -&72.5 & -\\
    ZSKD & 69.56 & 79.62 &-\\
    {Proxy Dataset} & 74.58 & 77.81 & 46.32\\
    DCGAN  & 66.24 & 79.67 & 39.77\\
    (Ours)  \textbf{DeGAN} & 80.55 & 83.79 & 65.25\\
    \hline
  \end{tabular}
  }
\end{table}
\section{Experiments}
\label{exp}
In this section, we discuss the experimental setup for an empirical evaluation of our proposed DeGAN framework. We use the benchmark datasets, CIFAR-$10$ \cite{krizhevsky2009learning}, Fashion-MNIST \cite{xiao2017fashion} and CIFAR-$100$ to demonstrate state-of-the-art results for the task of Data-Free Knowledge Distillation. We demonstrate the applicability of our approach to Class-Incremental Learning using CIFAR-$100$ dataset. We use PyTorch framework for all our implementations. 

\begin{table*}
\caption{Accuracy (in \%) of Student net with CIFAR-$10$ as \textit{True Dataset} (column headings indicate the \textit{Proxy Dataset} used) }

\centering
\resizebox{1\textwidth}{!}{
\begin{tabular}{l|ccc|ccccc|c|c}

\hline
  \multirow{2}{*}{Methods} & \multicolumn{3}{c|}{CIFAR-$100$ select samples} &
  \multicolumn{5}{c|}{\makecell{CIFAR-$100$ samples: 10 random classes per sample}} & 
  \multirow{2}{*}{\makecell{SVHN}} &
  \multirow{2}{*}{\makecell{Random \\ Noise}}\\
  \cline{2-9} 
  & \makecell{$90$ classes} & \makecell{$40$ classes} & \makecell{$6$ classes} & \makecell{Sample-1} & \makecell{Sample-2} & \makecell{Sample-3} & \makecell{Sample-4}  & \makecell{Sample-5} & {} \\ \hline\hline
  \textit{Proxy Data} & 74.58 & 65.78 & 36.44 & 42.15 & 49.24 & 46.65 & 49.08 & 47 & 45.18 & 11.63 \\
  DCGAN & 66.24 & 66.13 & 39.44 & 56.81 & 67.33 & 62.1 & 69.34 & 68.66 & 26.5 & 10.09 \\
  \textbf{DeGAN} (Ours) & \textbf{80.55} & \textbf{76.32} & \textbf{59.53} & \textbf{66.95} & \textbf{74.59} & \textbf{72.87} & \textbf{76.63} & \textbf{71.61} & \textbf{55.05} & \textbf{23.26} \\

\hline
\end{tabular}
}
\label{tab:char}
\end{table*}

\subsection{Knowledge Distillation Using DeGAN}
An overview of the experimental flow is presented below:
\begin{itemize}
    \item We first train a \textit{Teacher} model (or Classifier) on the \textit{True Dataset}. A train-validation split of $80$-$20$ is considered for this purpose. An early-stopping condition based on validation accuracy is set as the convergence criteria. The weights of the classifier are considered frozen for all future experiments.
    \item We use this trained classifier in the proposed DeGAN framework to construct representative samples of the \textit{True Dataset}. Samples from the \textit{Proxy Dataset} are used for training the DeGAN. For a given \textit{True Dataset}, we consider experiments with multiple \textit{Proxy Datasets}. We use the implementation of DCGAN from \citet{singh2019gan} as reference to implement the DeGAN. The learning rate for training the GAN is set to 0.0002 and a fixed number of epochs are trained ($200$ in all cases) to ensure consistency. The two hyper-parameters in this case are $\lambda_{e}$ and $\lambda_{d}$.
    \item The trained generator is used to perform the task of Knowledge Distillation. In every epoch of training, we generate a batch of samples using the generator. These samples are used to train the \textit{Student} using Knowledge Distillation(KD) loss \cite{hinton2015distilling}. We have given a weight of one to the distillation component of KD loss, and zero to the Cross-Entropy component. This is done to match conditions with the existing works \cite{nayak2019distillation}, and also to avoid additional hyper-parameter tuning. 
    \item We train a vanilla DCGAN in every case; and use this to perform KD to the \textit{Student} network. This serves as an ablation to prove the usefulness of the third element (Classifier) in DeGAN. The learning rate and number of training epochs are maintained same across the training of DCGAN and DeGAN. We also consider the baseline of using \textit{Proxy Data} directly for the task of Knowledge Distillation. This serves as a lower bound in each case. 
\end{itemize}

\subsubsection{Experiments with CIFAR-10 as \textit{True Dataset}}
CIFAR-$10$ \cite{krizhevsky2009learning} is a $10$-class labelled dataset consisting of colour images of size $32 \times 32$. This dataset has $50000$ training images and $10000$ test images. The images are equally distributed across all classes. We consider the \textit{Teacher} architecture as AlexNet and \textit{Student} architecture as AlexNet-half (network with half the capacity when compared to AlexNet), similar to that used by \citet{nayak2019distillation}. With CIFAR-$10$ as the \textit{True Dataset}, we consider the following \textit{Proxy Datasets}: CIFAR-$100$ select classes and SVHN. 
CIFAR-$100$ \cite{krizhevsky2009learning} consists of $100$ labelled classes with images of dimension $32 \times 32$. Each class has $500$ training images and $100$ test images. Hence the number of images per class in CIFAR-$100$ is one-tenth of that in CIFAR-$10$. The $100$ classes in this dataset can be grouped into $20$ categories. In this case, CIFAR-$100$ is a related dataset, since both CIFAR-$10$ and CIFAR-$100$ consist of natural images of the same size. Most of the images in both datasets belong to object classes, with a few exceptions in CIFAR-$100$. 
Although we consider a related domain dataset, we claim that the classes in the \textit{Proxy Dataset} can be unrelated to those in the \textit{True Dataset}. In order to demonstrate this, we consider multiple combinations of classes as \textit{Proxy Datasets}: (results in Table-\ref{tab:char})

\begin{enumerate}
    \item Only non-overlapping classes between CIFAR-$10$ and CIFAR-$100$ are used. The categories, vehicles1 and vehicles2 from CIFAR-$100$ are excluded here. ($90$ classes are used) 
    \item Categories in CIFAR-$100$ which are even remotely related to the CIFAR-$10$ classes have been excluded in this case. Categories used are: flowers, food containers, fruits and vegetables, household electric devices, household furniture, trees, large man-made outdoor things, large natural outdoor scenes (40 classes are used) 
    \item Only background classes are used in this case. The $6$ classes used are: Road, Cloud, Forest, Mountain, Plain and Sea. This case can be regarded as an unrelated dataset, since these are not object classes. So, the discriminator in the DeGAN does not learn the notion of an object from the \textit{Proxy Dataset} in this case. 
    \item 10 classes are randomly sampled from the $40$ handpicked unrelated classes and used as \textit{Proxy Dataset}. This random sampling process is repeated $5$ times. 
\end{enumerate}

In order to understand the true potential of our proposed approach, we consider the case when related datasets are not available. SVHN \cite{netzer2011reading} is a publicly available colour dataset consisting of street view house numbers cropped to the dimension $32 \times 32$. This dataset has $10$ classes, with each class representing one digit. This dataset consists of $73257$ training images and $26032$ test images. Since SVHN consists of digits, the statistics of images in this dataset will be different from that in the CIFAR-$10$ dataset. Hence this is a dataset from an unrelated domain. 

The consolidated results of all the above experiments are presented in Table-\ref{tab:char}. Our results have been compared with the state-of-the-art results in Table-\ref{cifar10-table}. We consider the case of CIFAR-$100$ with 90 classes as the \textit{Proxy Dataset} for this. The performance of our approach is better than the existing work \cite{nayak2019distillation} by $10.99\%$ as shown in Table-\ref{cifar10-table}. Although the total number of samples used in the $90$-class case is $45000$, which is lesser than the size of CIFAR-$10$ training data set, we are still able to closely match the performance of Knowledge Distillation using actual data samples. As we reduce the number of classes in the \textit{Proxy Dataset}, and as we move to classes that are more unrelated to the \textit{True Dataset}, the Knowledge Distillation (KD) accuracy drops. However, our approach is consistently better compared to the two baselines: directly using \textit{Proxy Data} for KD (denoted by \textit{Proxy Data} in Table-\ref{tab:char}); and using DCGAN to generate samples for KD.

We consider an ablation of using random noise as the \textit{Proxy Dataset}. While the baseline of using \textit{Proxy Dataset} directly gives an accuracy that is close to that of a random guess($11.63\%$), DeGAN is able to improve the accuracy significantly to $23.26\%$. The baseline of using DCGAN also gives the accuracy equivalent to a random guess ($10.09\%$). This experiment demonstrates the importance of enforcing a good prior on the generated images. This also demonstrates that the DeGAN framework can enrich \textit{any available Proxy Data} to make it more useful for a given task.

\subsubsection{Experiments with Fashion MNIST as \textit{True Dataset}}
Fashion MNIST \cite{xiao2017fashion} is a grayscale image dataset consisting of 10 object classes. The dimension of each image is $28 \times 28$. The dataset consists of $60000$ training samples, and $10000$ test samples. We consider the \textit{Teacher} architecture as LeNet \cite{lecun1998gradient} and \textit{Student} architecture as LeNet-half (network with half the capacity when compared to LeNet), similar to that used by \citet{nayak2019distillation}. We use Fashion MNIST as the \textit{True Dataset} and consider CIFAR-$10$ and SVHN as \textit{Proxy Datasets}. Both datasets (CIFAR-$10$ and SVHN) are converted to grayscale and used for training DeGAN. In this case, neither of these datasets belong to the domain of the \textit{True Dataset}. However, CIFAR-$10$ contains object classes, which is a property that even Fashion MNIST classes possess. Since SVHN has only numbers, it does not possess object-like features. So, CIFAR-$10$ is more related to the domain of Fashion MNIST dataset when compared to SVHN. We use \textit{Proxy Dataset} as CIFAR-$10$ for comparison with the state of the art (Table-\ref{cifar10-table}). We demonstrate that we can beat the performance of the existing approaches \cite{kimura2018few,nayak2019distillation} by at least $4.17\%$ for the task of Knowledge Distillation using CIFAR-$10$ dataset, although it is not related to the Fashion MNIST dataset. The margin with respect to the state-of-the-art approach is not as high as the previous case, as we did not consider a related dataset here.

\subsubsection{Experiments with CIFAR-100 as \textit{True Dataset}}
To demonstrate scalability of the proposed DeGAN framework, we use CIFAR-$100$ as the \textit{True Dataset}. We consider the \textit{Teacher} architecture as Inception-V3 \cite{szegedy2016rethinking,chollet2017xception} and \textit{Student} architecture as ResNet18 \cite{he2016deep}. We consider a related dataset, CIFAR-$10$ as the \textit{Proxy Dataset} here. The number of training data samples in CIFAR-$100$ is the same as that of CIFAR-$10$. However, the number of classes is much lesser in CIFAR-$10$. Although we consider a related dataset with 2 overlap classes (automobiles and trucks), this case is more challenging when compared to the above experiments since the number of classes in the \textit{True Dataset} is larger, and the number of classes in the \textit{Proxy Dataset} is much lesser. This may lead to a high class imbalance when Vanilla-GAN is used, resulting in a large number of misclassifications in the sparsely populated classes. The diversity loss in DeGAN helps maintain balance across the CIFAR-$100$ classes. The results are presented in Table-\ref{cifar10-table}. We see an improvement of about $19\%$ with respect to the baseline. This is a significant improvement for top $1\%$ accuracy of a 100-class dataset.

\subsection{Class-Incremental Learning Using DeGAN}
We consider the case of single-step Class-Incremental learning on CIFAR-$100$ dataset. This is done as proof of concept to support the claim that the data generated using DeGAN can be used to replace the \textit{True Dataset} for various tasks. 
An initial model is first trained on a random set of $20$ classes, which are termed as \textit{old classes}. The goal is to incrementally learn the next set of $20$ classes in a data-free setting, where the old class data is assumed to be unavailable. We use ResNet-32 \cite{he2016deep} architecture for the initial as well as final models, similar to the baselines we compare with \cite{rebuffi2017icarl}. We use the results reported by \citet{rebuffi2017icarl} for comparison. We consider the standard losses that are used in Incremental Learning \cite{li2017learning}: Cross-entropy loss to learn the \textit{new classes} and Distillation loss, to avoid catastrophic forgetting on the \textit{old classes}. In addition, we also add a regularization term to account for the relative scaling of \textit{logits} between the old and new classes. We use our proposed DeGAN to extract representative samples of \textit{old classes} using the \textit{new class} data as \textit{Proxy Data}. This generated data is used in the Distillation Loss component to avoid catastrophic forgetting of the \textit{old classes}. We compare our results with the baselines explained in Section \ref{rel_work}. We also include a baseline of using the \textit{new class} data directly in the Distillation Loss similar to LwF \cite{li2017learning}. The results in Table-\ref{incremental} demonstrate a significant improvement in the accuracy with respect to other Data-Free baselines. After one incremental update, since we have a trained generator, existing methods \cite{shin2017continual} that incrementally learn the generator and classifier sequentially, can be used for further updates. This could not be used for the first incremental update as this approach requires data from \textit{old classes} to train the first generator.
\begin{table}
  \caption{Single-step Class-Incremental Learning: Comparison with the state of the art\\}
  \label{incremental}
  \centering
  \begin{tabular}{l|c}
    \hline
    Methods & Accuracy (in \%) \\
    \hline
    Finetuning & 41.6 \\
    Fixed Representation & 46.8 \\
    LwF.MC & 62.58 \\
    Using \textit{Proxy Data} & 65.03 \\
    \textbf{DeGAN} (Ours) & \textbf{68.65}\\
    
    \hline
  \end{tabular}
\end{table}
\section{Conclusion}
\label{conc}
We have proposed a novel Data-enriching GAN (DeGAN) framework to enrich data from any domain, such that it is more suitable for the future tasks of a given trained classifier. The problem of retrieving representative samples from a trained classifier is of importance in several applications such as Knowledge Distillation, Incremental Learning, Visualization and Crafting of Adversarial Perturbations. We have empirically evaluated our framework on several benchmark datasets to demonstrate that we can achieve state-of-the-art results for the task of Data-Free Knowledge Distillation using data from a related domain. We observe that the samples generated using related domain data can also serve as useful visualizations for the \textit{True Dataset}. 

We have further demonstrated that the data generated using DeGAN can replace the training dataset for multiple tasks, and hence is truly representative of the same. We show state-of-the-art results on the task of Class-Incremental learning, where we do not have access to old class data. 

Since it is not easy to quantitatively judge how similar the domain of the \textit{Proxy Dataset} is, we also evaluate the impact of using unrelated domain data as the \textit{Proxy Dataset}. We show that even if the \textit{Proxy Data} is unrelated to the \textit{True Data}, our proposed DeGAN can significantly enrich the dataset such that it is more useful than using vanilla-GAN for the future tasks.
\subsubsection{Acknowledgements}
This work was partially supported by the Robert Bosch Center for Cyber-Physical Systems (RBCCPS), Indian Institute of Science (IISc). We would also like to extend our gratitude to the students and research assistants at Video Analytics Lab, IISc for the insightful discussions on this work.

\bibliography{DeGAN_arxiv}
\bibliographystyle{aaai}

\end{document}